\documentclass{article}

    \usepackage[final]{tackling_climate_workshop_style}

\usepackage[utf8]{inputenc} 
\usepackage[T1]{fontenc}    
\usepackage{hyperref}       
\usepackage{url}            
\usepackage{booktabs}       
\usepackage{amsfonts}       
\usepackage{nicefrac}       
\usepackage{microtype}      

\usepackage{subcaption}
\usepackage{multirow}
\usepackage{amsmath}
\usepackage{amssymb}
\usepackage{mathtools}
\usepackage{amsthm}
\usepackage{natbib}
\usepackage[capitalize,noabbrev]{cleveref}

\usepackage{wrapfig}
\theoremstyle{plain}

\theoremstyle{definition}

\theoremstyle{remark}

\title{Multiscale Neural PDE Surrogates for Prediction and Downscaling: Application to Ocean Currents}

\author{%
  Abdessamad El Kabid\thanks{Corresponding author.} \\
  Mila -- Qu\'ebec AI Institute \\
  \'Ecole polytechnique \\
  Palaiseau, France \\
  \texttt{abdessamad.el-kabid@polytechnique.edu} \\
  \And
  Loubna Benabbou \\
  Mila -- Qu\'ebec AI Institute \\
  Universit\'e du Qu\'ebec \`a Rimouski \\
  L\'evis, Qu\'ebec, Canada \\
  \And
  Redouane Lguensat \\
  Institut Pierre-Simon Laplace, IRD \\
  Sorbonne Universit\'e \\
  Paris, France \\
  \And
  Alex Hern\'andez-Garc\'ia \\
  Mila -- Qu\'ebec AI Institute \\
  Universit\'e de Montr\'eal \\
  Montr\'eal, Qu\'ebec, Canada \\
}

\begin{document}

\maketitle

\begin{abstract}
Accurate modeling of physical systems governed by partial differential equations is a central challenge in scientific computing. In oceanography, high-resolution current data are critical for coastal management, environmental monitoring, and maritime safety. However, widely used satellite products, such as Copernicus sea-surface velocity at $\sim 0.08^\circ$ resolution and global ocean models, often lack the spatial granularity required for detailed local analyses. We (a) introduce a supervised deep learning framework based on neural operators for solving PDEs and producing arbitrary-resolution solutions, and (b) propose downscaling models applied to Copernicus ocean current data. Additionally, our method serves as a surrogate PDE model that predicts solutions at arbitrary resolution, regardless of the input resolution. We evaluate on real-world Copernicus ocean current data and synthetic Navier--Stokes simulation datasets.
\end{abstract}

\section{Introduction}

Accurate and high-resolution marine current fields are foundational to numerous marine applications, coastal engineering design, and autonomous navigation. Datasets like Copernicus ocean analysis product \cite{CMEMS_GLO_PHY_2025} provide global coverage at roughly $0.08^\circ \times 0.08^\circ$ (approximately 9 km in mid-latitudes), which are insufficient for applications requiring detailed local dynamics.

Downscaling methods, both dynamical and statistical, have been used to bridge this resolution gap. While dynamical downscaling, employing regional ocean models, is physically rigorous, it demands substantial computational resources, often requiring days of runtime on HPC clusters. Statistical approaches offer computational efficiency but struggle with the multiscale and non-linear nature of fluid dynamics~\cite{kruyt2022downscaling}. 

Deep learning has emerged as an alternative to traditional statistical methods, since it can learn from data the mappings from coarse- to fine-scale representations. Initial efforts with CNNs and GANs achieved success in meteorology \citep{vosper2023} and image-based super‑resolution \citep{dong2015}. However, these models typically require fixed upsampling factors and lack fidelity when generalizing to unseen resolutions or evolving physical dynamics.

Neural operators, particularly Fourier Neural Operators (FNOs) \citep{li2020fourier}, Transolver \citep{wu2024transolver}, FactFormer \citep{sun2024factformer}, and Latent Spectral Models (LSMs) \citep{wu2023solvinghighdimensionalpdeslatent}, have demonstrated remarkable ability to learn operators governing PDEs. FNOs have been successfully applied to atmospheric and oceanographic forecasting \citep{sun2024}.


\citet{yang2023fourierneuraloperatorsarbitrary}'s DFNO model addressed downscaling for climate data and PDE solutions at arbitrary resolutions, where the PDE solution is generated via numerical solvers, not the model itself. Our work extends this paradigm in two significant directions. First, we generalize their model to handle temporal sequences, enabling the prediction of PDE solutions at arbitrary spatial resolutions using the same model. Second, we benchmark multiple downscaling models inspired by the DFNO by applying it to the Copernicus ocean current dataset for static downscaling, to demonstrate its real-world impact geophysical data.

Our main contributions are as follows.
\begin{itemize}
\item We benchmark multiple models for arbitrary-resolution downscaling and apply them to physical observations that need downscaling (ocean current from Copernicus marine data).
\item We develop a surrogate model capable of predicting PDE solutions at arbitrary resolutions-independent of the input resolution, giving more flexibility and extent to the model.
\end{itemize}

We present, to our knowledge, one of the first PDE surrogate approaches that generates solutions at arbitrary output resolution independent of the input resolution

\section{Methodology}

\subsection{Resolution-agnostic Neural Operator Framework for Downscaling}

\begin{figure*}[h]
  \centering
  \includegraphics[width=\linewidth]{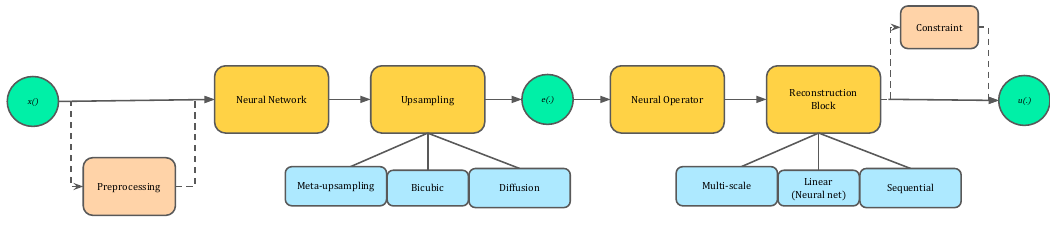}
 \caption{The figure (inspired by \citet{yang2023fourierneuraloperatorsarbitrary}) shows the overall structure of our Temporal/static downscaling model. The low-resolution input a goes through an optional preprocessing (gradient transformation for gradient based methods) then a neural network and an upsampling block. Then an embedding function $e(·)$ is returned. Finally, a neural operator takes in $e(·)$ and outputs a function which gets into a reconstruction block and an optional constraint layer.}
  \label{fig:Framework}
\end{figure*}

We propose a flexible framework for simultaneous PDE solution prediction and downscaling, applicable to both temporal and static settings. The general architecture, illustrated in Figure~\ref{fig:Framework}, begins with a low-resolution input that optionally undergoes preprocessing—such as gradient transformation in gradient-based models—followed by a neural network and an upsampling block. The resulting output is then processed by a neural operator. Finally, a reconstruction block and an optional physical constraint layer complete the pipeline.

For the specific task of ocean current downscaling, we evaluated several model variants within this framework. We first introduced \textbf{DUNO}, which uses a U-shaped Neural Operator (UNO) in the neural operator block. UNO generalizes well across different PDE types and is more expressive than the standard FNO. the rest of the framework follows DFNO \cite{yang2023fourierneuraloperatorsarbitrary}.

We also introduce the \textbf{SpecDFNO} which extends the standard DFNO architecture by introducing a second neural operator that predicts the residual between the initial FNO output and the ground truth. Inspired by \citet{qin2024betterunderstandingfourierneural}, this residual is then added to the base prediction, enhancing the model's ability to capture high-frequency components often lost in downscaling.

Extending this further, the \textbf{SpecDFNO with Diffusion-Based Upsampling (SpecDFNODiff)} replaces the explicit upsampling operation with a learned generative diffusion prior. The diffusion process is conditioned on the low-resolution input, allowing the model to generate high-resolution fields that are spatially coherent and physically plausible.

We also explored gradient-based strategies. The \textbf{MetaGradDFNO} model applies the DFNO architecture on gradient fields derived using Sobel filters (applied in the preprocessing block). This model also uses a meta-learning mechanism (in the upsampling block) that learns a weighted combination of nearest-neighbor, bilinear, and bicubic interpolation kernels, enabling context-aware upsampling.

Complementary to this, the \textbf{Multiscale Gradient DFNO (MultiGradDFNO)} captures structural information across multiple scales by using parallel convolutional branches with varying kernel sizes to process gradient fields (in the reconstruction block). These branches are then merged via a convolution, which helps retain rich spatial features present even at coarse resolution.

Across all models, physical conservation laws are softly enforced through a softmax constraint layer, a mechanism demonstrated to be effective in geophysical settings by \citet{harder2024hardconstraineddeeplearningclimate}.

\subsection{Models for Direct PDE Solution Prediction at Multiple Scales}

In addition to downscaling tasks, we adapt our architecture for direct prediction of PDE solutions directly at multiple resolutions. The models \textbf{Temp\_DFNO} and \textbf{Temp\_SpecDFNO} extend the DFNO and SpecDFNO architectures by incorporating a temporal dimension. In these versions, the convolutions are performed not only across spatial axes but also along the temporal axis, allowing the network to capture spatiotemporal dynamics inherent in time-evolving PDE systems.

\section{Experiments}

\subsection{Datasets and Training}
The datasets, training configuration and description of the baselines can be found in Appendix~\ref{app:training}.
\subsection{Temporal Models for Multiscale PDE Solving}

We first evaluate the performance of the model on the 2D incompressible Navier-Stokes dataset. The goal is to learn a surrogate that accurately simulates spatio-temporal dynamics across multiple spatial resolutions~(Appendix \ref{app:visualizations}). Results for two DFNO variants—standard (\textbf{Temp\_DFNO}) and residual-based (\textbf{Temp\_SpecDFNO})—are reported in Table~\ref{tab:temporal_dfno_results}.



Our temporal models deliver accurate predictions across multiple spatial resolutions on the Navier–Stokes benchmark. They achieve competitive performance compared to DFNO-2/-4 \citet{yang2023fourierneuraloperatorsarbitrary} while enabling multi-resolution prediction within a single network. Unlike DFNO-2/-4, which \emph{depend on an external numerical solver} to generate coarse solutions prior to downscaling, our approach unifies solution generation and resolution-agnostic inference in a single framework. This combines the predictive strengths associated with classical solvers and the physics-aware downscaling of DFNO, without requiring precomputed low-resolution solutions.




\begin{table}[htb!]
\small
\centering
\caption{Performance of Temporal\_DFNO, Temporal\_SpecDFNO, and DFNO models at different resolutions. DFNO-2 and DFNO-4 values are taken from \cite{yang2023fourierneuraloperatorsarbitrary}; PSNR and SSIM for these models were not provided.\\} 
\begin{tabular}{llcccc}
\hline
\textbf{Model} & \textbf{Resolution} & \textbf{MAE} & \textbf{MSE} & \textbf{PSNR} & \textbf{SSIM} \\
\hline
\multirow{3}{*}{Temp\_DFNO} 
& 16x16 & 0.017941 & 0.000603 & 33.82 & 0.9920 \\
& 32x32 & 0.017426 & 0.000573 & 34.05 & 0.9878 \\
& 64x64 & 0.019775 & 0.000722 & 32.64 & 0.9805 \\
\hline
\multirow{3}{*}{Temp\_SpecDFNO} 
& 16x16 & 0.017806 & 0.000599 & 34.00 & 0.9915 \\
& 32x32 & 0.017372 & 0.000568 & 34.15 & 0.9882 \\
& 64x64 & 0.019736 & 0.000712 & 32.76 & 0.9811 \\
\hline
\multirow{2}{*}{DFNO-2 } 
& 32x32 & 0.0124 & 0.0004 & -- & -- \\
& 64x64 & 0.0246 & 0.0018 & -- & -- \\
\hline
\multirow{2}{*}{DFNO-4 } 
& 32x32 & 0.0208 & 0.0012 & -- & -- \\
& 64x64 & 0.0168 & 0.0007 & -- & -- \\
\hline
\end{tabular}
\label{tab:temporal_dfno_results}
\end{table}

\subsection{Downscaling Copernicus ocean current data}

    


To evaluate the performance of our models on real-world oceanographic data, we used sea surface velocity fields provided by the Copernicus Marine Environment Monitoring Service. To simulate coarse observations, we downsampled the original fields using average pooling to generate inputs at resolutions corresponding to 2×, 4×, and 8× coarsening factors ~(Appendix \ref{app:visualizations}). The model was then tasked with reconstructing higher-resolution fields from the lowest-resolution inputs. Unlike the synthetic Navier–Stokes dataset, no temporal supervision was used; each sample was treated independently as a static snapshot. 

The proposed DFNO variants significantly outperform conventional CNN baselines at all downscaling levels. However, as the resolution gap increases (at 8×), performance deteriorates (Table~\ref{tab:downscale-128}) due to the lack of informative coarse-scale details. 

    \paragraph{2× downscaling} (16×16 → 32×32): All neural operator variants substantially outperform CNN baselines. MetaGradDFNO achieves the best overall performance. The dramatic performance gap between neural operators and CNNs (MAE improvement of ~30×) demonstrates the importance of spectral representations for fluid flow reconstruction.

    \paragraph{4× downscaling }(16×16 → 64×64): Performance degradation becomes evident as the reconstruction task becomes more challenging. SpecDFNO emerges as the most robust and consistent model, while gradient-based variants show reduced effectiveness at this scale and later scales. Interestingly, DUNO maintains competitive MSE performance.

    \paragraph{8× downscaling}(16×16 → 128×128)(Appendix~\ref{app:limitations}) : Significant performance degradation occurs in all models. DUNO demonstrates good performance, and SpecDFNO and SpecDFNODiff are overall and perceptually better.

\begin{table}[t]
\centering
\scriptsize
\caption{Downscaling results at $32\times32$ and $64\times64$ .}
\label{tab:downscale-32-64-one}
\begin{tabular}{llcccc|cccc}
\toprule
\multicolumn{2}{c}{} & \multicolumn{4}{c}{\textbf{32$\times$32}} & \multicolumn{4}{c}{\textbf{64$\times$64}} \\
\cmidrule(lr){3-6}\cmidrule(lr){7-10}
\textbf{Model} & \textbf{Loss} & \textbf{MAE} & \textbf{MSE} & \textbf{PSNR} & \textbf{SSIM} & \textbf{MAE} & \textbf{MSE} & \textbf{PSNR} & \textbf{SSIM} \\
\midrule
\multirow{2}{*}{CNN\_2x} & L1 & 0.424  & 0.19343 & 14.06813 & 0.10542 & 0.4246 & 0.1945 & 15.3113 & 0.0992 \\
                         & L2 & 0.424  & 0.18774 & 14.19797 & 0.12992 & 0.4244 & 0.1890 & 15.4370 & 0.1123 \\
\addlinespace
\multirow{2}{*}{CNN\_4x} & L1 & 0.3954 & 0.16950 & 14.64186 & 0.12584 & 0.3953 & 0.1720 & 15.8440 & 0.1210 \\
                         & L2 & 0.395  & 0.16639 & 14.72229 & 0.12872 & 0.3954 & 0.1663 & 15.9890 & 0.1163 \\
\addlinespace
\multirow{2}{*}{DFNO}    & L1 & 0.02134 & 0.00099 & 36.97443 & 0.96785 & 0.04057 & 0.01749 & 33.15377 & 0.85038 \\
                         & L2 & 0.01822 & 0.00059 & 39.23959 & 0.97699 & 0.04151 & 0.01640 & 33.23573 & 0.84019 \\
\addlinespace
\multirow{2}{*}{DUNO}    & L1 & 0.03501 & 0.00254 & 32.88501 & 0.91708 & 0.04271 & \underline{0.00320} & 32.18915 & 0.82181 \\
                         & L2 & 0.03440 & 0.00217 & 33.57706 & 0.92083 & 0.04251 & \textbf{0.00314} & 32.55057 & 0.82200 \\
\addlinespace
\multirow{2}{*}{MetaGradDFNO} & L1 & 0.01424 & 0.00045 & 40.37440 & 0.98574 & 0.05493 & 0.00962 & 30.33814 & 0.73601 \\
                              & L2 & \textbf{0.01370} & \textbf{0.00033} & \textbf{41.74495} & \underline{0.98638} & 0.05685 & 0.00851 & 30.10898 & 0.71987 \\
\addlinespace
\multirow{2}{*}{MultiGradDFNO} & L1 & 0.01465 & 0.00049 & 40.07315 & 0.98396 & 0.05513 & 0.00851 & 30.04188 & 0.73671 \\
                               & L2 & \underline{0.01380} & \underline{0.00034} & \underline{41.58843} & \textbf{0.98690} & 0.05672 & 0.00802 & 29.83308 & 0.71801 \\
\addlinespace
\multirow{2}{*}{SpecDFNO} & L1 & 0.03407 & 0.00242 & 33.09110 & 0.92286 & 0.04283 & 0.00682 & 32.23081 & 0.82589 \\
                          & L2 & 0.02451 & 0.00106 & 36.66621 & 0.95871 & \textbf{0.03653} & 0.00655 & \textbf{34.16133} & \textbf{0.87433} \\
\addlinespace
\multirow{2}{*}{SpecDFNODiff} & L1 & 0.01836 & 0.00064 & 38.85981 & 0.97519 & \underline{0.03796} & 0.00367 & \underline{33.97083} & \underline{0.86536} \\
                              & L2 & 0.01712 & 0.00049 & 40.07214 & 0.97818 & 0.03879 & 0.00355 & 33.90497 & 0.86177 \\
\bottomrule
\end{tabular}
\end{table}

\section{Conclusion}
We show that neural-operator downscaling can deliver higher-resolution current maps from low-resolution inputs, with strong gains over CNN baselines for Copernicus currents and \textit{accurate multi-resolution temporal predictions without external solvers}. These results position neural operators as practical, scalable tools for ocean current analysis, and motivate uncertainty-aware and theory-driven extensions to safely push beyond moderate downscaling.




%

\clearpage

\bibliography{tackling_climate_workshop}

\begin{thebibliography}{16}
\providecommand{\natexlab}[1]{#1}
\providecommand{\url}[1]{\texttt{#1}}
\expandafter\ifx\csname urlstyle\endcsname\relax
  \providecommand{\doi}[1]{doi: #1}\else
  \providecommand{\doi}{doi: \begingroup \urlstyle{rm}\Url}\fi

\bibitem[Beucler et~al.(2021)Beucler, Pritchard, Rasp, Ott, Baldi, and Gentine]{beucler2021}
Beucler, T., Pritchard, M., Rasp, S., Ott, J., Baldi, P., and Gentine, P.
\newblock Enforcing analytic constraints in neural networks emulating physical systems.
\newblock \emph{Physical Review Letters}, 126\penalty0 (9):\penalty0 098302, 2021.

\bibitem[Campbell et~al.(2025)Campbell, Warder, Bhaskaran, and Piggott]{CAMPBELL2025100485}
Campbell, A.~M., Warder, S.~C., Bhaskaran, B., and Piggott, M.~D.
\newblock Domain-informed cnn architectures for downscaling regional wind forecasts.
\newblock \emph{Energy and AI}, 20:\penalty0 100485, 2025.
\newblock ISSN 2666-5468.
\newblock \doi{https://doi.org/10.1016/j.egyai.2025.100485}.
\newblock URL \url{https://www.sciencedirect.com/science/article/pii/S2666546825000175}.

\bibitem[Chattopadhyay(2023)]{chattopadhyay2023}
Chattopadhyay, S. e.~a.
\newblock Oceannet: Surrogate modeling of gulf stream dynamics using fno and predictor–corrector schemes.
\newblock \emph{Advances in Water Resources}, 2023.

\bibitem[{Copernicus Marine Service}(2025)]{CMEMS_GLO_PHY_2025}
{Copernicus Marine Service}.
\newblock Global ocean physics analysis and forecast (cmems).
\newblock \url{https://doi.org/10.48670/moi-00016}, 2025.
\newblock Accessed: 2025-06-11.

\bibitem[Daw(2020)]{daw2020}
Daw, D. B. e.~a.
\newblock Physics-guided architecture (pga) of neural networks for quantifying uncertainty in lake temperature modeling.
\newblock \emph{SIAM Journal on Scientific Computing}, 42\penalty0 (3):\penalty0 1450--1470, 2020.

\bibitem[Dong(2015)]{dong2015}
Dong, C. e.~a.
\newblock Image super-resolution using deep convolutional networks.
\newblock In \emph{IEEE Transactions on Pattern Analysis and Machine Intelligence}, volume~38, pp.\  295--307, 2015.

\bibitem[Harder et~al.(2024)Harder, Hernandez-Garcia, Ramesh, Yang, Sattigeri, Szwarcman, Watson, and Rolnick]{harder2024hardconstraineddeeplearningclimate}
Harder, P., Hernandez-Garcia, A., Ramesh, V., Yang, Q., Sattigeri, P., Szwarcman, D., Watson, C., and Rolnick, D.
\newblock Hard-constrained deep learning for climate downscaling, 2024.
\newblock URL \url{https://arxiv.org/abs/2208.05424}.

\bibitem[Kruyt et~al.(2022)Kruyt, Mott, Fiddes, Gerber, Sharma, and Reynolds]{kruyt2022downscaling}
Kruyt, B., Mott, R., Fiddes, J., Gerber, F., Sharma, V., and Reynolds, D.
\newblock A downscaling intercomparison study: The representation of slope- and ridge-scale processes in models of different complexity.
\newblock \emph{Frontiers in Earth Science}, 10:\penalty0 789332, 2022.
\newblock \doi{10.3389/feart.2022.789332}.

\bibitem[Li et~al.(2020)Li, Kovachki, Azizzadenesheli, Liu, Bhattacharya, Stuart, and Anandkumar]{li2020fourier}
Li, Z., Kovachki, N., Azizzadenesheli, K., Liu, B., Bhattacharya, K., Stuart, A., and Anandkumar, A.
\newblock Fourier neural operator for parametric partial differential equations.
\newblock \emph{arXiv}, abs/2010.08895, 2020.

\bibitem[Li et~al.(2023)Li, Shu, and Farimani]{sun2024factformer}
Li, Z., Shu, D., and Farimani, A.~B.
\newblock Scalable transformer for pde surrogate modeling, 2023.
\newblock URL \url{https://arxiv.org/abs/2305.17560}.

\bibitem[Qin et~al.(2024)Qin, Lyu, Peng, Geng, Wang, Tang, Leroyer, Gao, Liu, and Wang]{qin2024betterunderstandingfourierneural}
Qin, S., Lyu, F., Peng, W., Geng, D., Wang, J., Tang, X., Leroyer, S., Gao, N., Liu, X., and Wang, L.~L.
\newblock Toward a better understanding of fourier neural operators from a spectral perspective, 2024.
\newblock URL \url{https://arxiv.org/abs/2404.07200}.

\bibitem[Sun et~al.(2024)Sun, Sowunmi, Egele, Narayanan, Roekel, and Balaprakash]{sun2024}
Sun, Y., Sowunmi, O., Egele, R., Narayanan, S. H.~K., Roekel, L.~V., and Balaprakash, P.
\newblock Streamlining ocean dynamics modeling with fourier neural operators: A multiobjective hyperparameter and architecture optimization approach, 2024.
\newblock URL \url{https://arxiv.org/abs/2404.05768}.

\bibitem[Vosper et~al.(2022)Vosper, Harris, McRae, Aitchison, Watson, Santos~Rodriguez, and Mitchell]{vosper2023}
Vosper, E., Harris, L., McRae, A., Aitchison, L., Watson, P., Santos~Rodriguez, R., and Mitchell, D.
\newblock Deep learning for downscaling tropical cyclone rainfall.
\newblock In \emph{NeurIPS 2022 Workshop on Tackling Climate Change with Machine Learning}, 2022.
\newblock URL \url{https://www.climatechange.ai/papers/neurips2022/13}.

\bibitem[Wu et~al.(2023)Wu, Hu, Luo, Wang, and Long]{wu2023solvinghighdimensionalpdeslatent}
Wu, H., Hu, T., Luo, H., Wang, J., and Long, M.
\newblock Solving high-dimensional pdes with latent spectral models, 2023.
\newblock URL \url{https://arxiv.org/abs/2301.12664}.

\bibitem[Wu et~al.(2024)Wu, Luo, Wang, Wang, and Long]{wu2024transolver}
Wu, H., Luo, H., Wang, H., Wang, J., and Long, M.
\newblock Transolver: A fast transformer solver for pdes on general geometries.
\newblock In \emph{Proceedings of the 41st International Conference on Machine Learning (ICML)}, 2024.

\bibitem[Yang et~al.(2023)Yang, Hernandez-Garcia, Harder, Ramesh, Sattegeri, Szwarcman, Watson, and Rolnick]{yang2023fourierneuraloperatorsarbitrary}
Yang, Q., Hernandez-Garcia, A., Harder, P., Ramesh, V., Sattegeri, P., Szwarcman, D., Watson, C.~D., and Rolnick, D.
\newblock Fourier neural operators for arbitrary resolution climate data downscaling, 2023.
\newblock URL \url{https://arxiv.org/abs/2305.14452}.

\end{thebibliography}
\bibliographystyle{icml2025}

\newpage
\appendix
\section{Related Work}

\subsection{Neural Operators}

Neural operators are models that learn mappings between infinite-dimensional function spaces and have recently emerged as powerful tools to approximate solution operators of partial differential equations (PDEs)~\citep{li2020fourier, wu2024transolver, sun2024factformer}. These models achieve strong performance on a variety of parametric PDE benchmarks, including Navier-Stokes, the Darcy flow, and Burgers' equation, while offering orders-of-magnitude speed-ups over classical numerical solvers. The Fourier Neural Operator (FNO)~\citep{li2020fourier} performs operator learning in Fourier space, enabling efficient global convolution. Latent Spectral Models (LSMs) project high-dimensional PDE fields into lower-dimensional latent spaces, where the equations are solved to improve both accuracy and computational efficiency for fluid and solid mechanics~\citep{wu2023solvinghighdimensionalpdeslatent}. Transolver~\citep{wu2024transolver} introduces physics-informed attention mechanisms, enabling the learning of PDE dynamics on unstructured meshes and complex geometries, thus reducing discretization dependence and surpassing previous neural operator architectures.

Neural operators are increasingly applied in oceanography. For example, \citet{chattopadhyay2023} proposed OceanNet, a hybrid FNO and predictor–evaluate–corrector model that learns Gulf Stream circulation dynamics and achieves up to $5 \times 10^5$ times speedup over classical numerical ocean models.

\subsection{Embedding Physics in Deep Learning}

In physics-based applications, it is critical that neural network outputs not only approximate ground truth but also remain consistent with the governing physical laws, which is essential for downstream applications and model trustworthiness. Incorporating physical priors into neural models has been shown to better capture observed physical properties. Techniques such as soft and hard constraint losses have been applied in atmospheric emulation, where physics-constrained models achieve lower errors while maintaining fidelity to the underlying equations~\citep{beucler2021, daw2020}. Moreover, \citet{yang2023fourierneuraloperatorsarbitrary} demonstrated that the introduction of physics-informed constraint layers further enhances fidelity and reduces error in climate downscaling tasks~\citep{yang2023fourierneuraloperatorsarbitrary, harder2024hardconstraineddeeplearningclimate}.

\subsection{Arbitrary-Resolution Downscaling}

Conventional neural network downscaling models, which operate between finite-dimensional spaces, are typically limited to fixed input and output sizes. As a result, a single trained model can only downscale inputs with a predefined upsampling factor; that is, the output resolution must match the resolution anticipated during training. For example, CNN-based methods have been used to downscale meteorological fields such as wind~\citep{CAMPBELL2025100485}, precipitation~\citep{vosper2023}, and solar radiation, often employing multistep cascades to achieve high-resolution output. However, these models exhibit degraded performance when applied to unseen upsampling factors.

To address this limitation, \citet{yang2023fourierneuraloperatorsarbitrary} introduced an FNO-based zero-shot downscaling model that generalizes to arbitrary resolutions without retraining. This approach outperforms both traditional super-resolution models and conventional neural PDE solvers on Navier–Stokes simulations and ERA5 climate fields. However, it is important to note that, in the PDE setting, the model only performs downscaling on solutions generated by external numerical solvers. \textit{ We overcome this limitation by developing a neural operator that (a) directly solves PDEs and (b) generates solutions at arbitrary resolution, independent of input resolution.
}

\section{Training configuration}
\label{app:training}
\subsection{Data Sources}
To evaluate our proposed model, we have considered as data sources Navier-Stokes data and satellite observations of the ocean currents.
\paragraph{Navier-Stokes}
    We used synthetic velocity fields based on the 2D incompressible Navier–Stokes equations in vorticity form on the periodic unit torus $\Omega = (0,1)^2$: 

\begin{align}
\partial_t \omega(x,t) + \mathbf{u}(x,t) \cdot \nabla \omega(x,t) &= \nu \Delta \omega(x,t) + f(x), \label{eq:ns-eq} \\
\nabla \cdot \mathbf{u}(x,t) &= 0, \\
\omega(x,0) &= \omega_0(x),
\end{align}

where $\omega$ is the scalar vorticity, $\mathbf{u}$ is the velocity field, and $\nu = 10^{-4}$ is the viscosity. The velocity is recovered from the vorticity via the stream function $\psi$, using:

\begin{equation}
\mathbf{u} = (\partial_y \psi, -\partial_x \psi), \quad -\Delta \psi = \omega.
\end{equation}

Following \cite{li2020fourier}, the forcing term is fixed as $f(x) = 0.1 \left( \sin(2\pi(x_1 + x_2)) + \cos(2\pi(x_1 + x_2)) \right)$, and the initial vorticity $\omega_0(x)$ is sampled from a Gaussian random field with mean zero and spectral decay: $\omega_0 \sim \mathcal{N}(0, r^{3/2} (-\Delta + 49I)^{-2.5})$.  

    A total of 10,000 simulations were run at a spatial resolution of $64 \times 64$, using randomly initialized conditions. Each simulation was evolved over 50 time steps with a fixed viscosity of $10^{-4}$. The dataset was split into 7,000 training samples, 2,000 validation samples, and 1,000 test samples. For each time step, we also constructed lower-resolution versions of the data by applying average pooling to obtain $32 \times 32$ and $16 \times 16$ grids, we gave 5 time steps as input and predicted the next 5 timesteps. The final dataset contains both the full-resolution solutions and their down sampled counterparts from the same timesteps and a window of 5 consequent timesteps.

\paragraph{Copernicus Data} 
    Real-world ocean current data was obtained from the Copernicus Marine Environment Monitoring Service (CMEMS). The dataset consists of global ocean surface velocities at 0.08° spatial resolution ($\sim$8 km), providing northward and eastward velocity components. This data combines satellite altimetry, in situ observations, and numerical ocean models through data assimilation. We selected regional subsets covering different oceanographic regimes to evaluate the generalization of the model in varying flow characteristics and coastal dynamics. We split the data into $128 \times 128$ patches. The dataset was then divided into 800 training samples, 200 validation samples, and 100 test samples.

\subsection{Data Preprocessing}
\label{app:preprocessing}

\begin{itemize}
    \item \textbf{Normalization:} Z-score normalization is applied per velocity component (northward and eastward) independently.
    \item \textbf{Train/Validation/Test Split:} The datasets are split using a 70\%/15\%/15\% ratio.

\end{itemize}

\subsection{Training Strategy}

\paragraph{Temporal Modeling (for PDE Surrogate Task):}
The model receives a sequence of five consecutive low resolution frames ($16 \times 16$) as input and predicts the next five frames at both low ($16 \times 16$) and high ($32 \times 32$) resolutions. The zero shot evaluation is performed on the $64 \times 64$ output from the $16 \times 16$ input.

\paragraph{Benchmarking Static Downscaling on Copernicus Data:}
Models are evaluated on the task of static downscaling using Copernicus current marine data. Inputs consist of low-resolution velocity fields, and models predict high-resolution outputs ($2\times$, $4\times$ and $8\times$ downscaling). Regional subsets representing different oceanographic regimes are used to assess generalization.

As baselines, we used CNNx2 and CNNx4 models. Each is implemented as a four-level U-Net, with increasing feature dimensions at each level (64, 128, 256, 512). Each encoder level employs a double convolution block comprising two 3×3 convolutional layers with batch normalization and ReLU activation, followed by 2×2 max pooling for spatial downsampling. The decoder mirrors the encoder structure with transposed convolutions for upsampling and skip connections to preserve fine-grained spatial information. CNNx2 and CNNx4 refer to training with 2× and 4× downsampling, respectively. For evaluation on both 2 times and 4 times downscaling. The 2 times downscaling outputs by CNN-2 increase their resolution to 4 times through model recursion and bicubic interpolation. The 4 times downscaling outputs by CNN-4 decrease their resolution to 2 times through average pooling and bicubic interpolation.

\paragraph{Loss Functions and Normalization:} We employ both L1 and L2 losses during training and evaluation. For perceptual quality assessment, PSNR and SSIM metrics are also computed (details provided in the Appendix). Additionally, input channels are normalized independently using channel-wise normalization.

\subsection{Optimizer and Hyperparameters.}
We train all models using the Adam optimizer with an initial learning rate of $1 \times 10^{-3}$. Each model is trained for 600 epochs with a batch size of 16.

\section{Limitations}
\label{app:limitations}
    \paragraph{Gradient-Enhanced Models:} MetaGradDFNO and MultiGradDFNO excel at moderate downscaling factors, leveraging multiscale information. However, their performance diminishes at higher ratios, where structural details become increasingly sparse.
    \paragraph{Spectral Residual Methods:} SpecDFNO shows robust performance across all scales, particularly excelling at zero-shot 4× downscaling. The diffusion-enhanced variant (SpecDFNODiff) provides marginal improvements but with an increased computational overhead.

\paragraph{Zero-shot 8 times downscaling}

\begin{table}[h]
\scriptsize
\centering
\caption{Downscaling to $128\times128$}
\label{tab:downscale-128}
\begin{tabular}{llcccc}
\toprule
Model & Loss & MAE & MSE & PSNR & SSIM \\
\midrule
CNN\_2x & L1 & 0.424 & 0.1952 & 15.8162 & 0.09986\\
 & L2 & 0.424 & 0.1895 & 15.944 & 0.105 \\
CNN\_4x & L1 & 0.395 & 0.1723 & 16.358 & 0.1176 \\
 & L2 & 0.395 & 0.1666 & 16.504 & 0.111\\
DFNO & L1 & 0.06488 & 0.00936 & 29.00970 & 0.60895 \\
 & L2 & 0.06488 & 0.00936 & 29.00905 & 0.60892 \\
DUNO & L1 & 0.06199 & 0.00797 & 29.70524 & 0.62467 \\
 & L2 & \textbf{0.05851} & 0.00755 & 29.94049 & \textbf{0.65233} \\
MetaGradDFNO & L1 & 0.06477 & 0.00934 & 29.01829 & 0.60980 \\
 & L2 & 0.06479 & 0.00935 & 29.01331 & 0.60976 \\
MultiGradDFNO & L1 & 0.06488 & 0.00894 & 29.20734 & 0.60906 \\
 & L2 & 0.06486 & 0.00936 & 29.00910 & 0.60886 \\
SpecDFNO & L1 & 0.06443 & 0.00923 & 29.06780 & 0.61219 \\
 & L2 & \underline{0.05971} & \textbf{0.00748} & \textbf{29.98298} & \underline{0.64175} \\
SpecDFNODiff & L1 & 0.06052 & \underline{0.00752} & \underline{29.96232} & 0.63049 \\
 & L2 & 0.06386 & 0.00788 & 29.75918 & 0.60965 \\
\bottomrule
\end{tabular}
\end{table}

While our models show robust performance at moderate resolution increases (2× and 4×), their accuracy degrades with higher downscaling factors (e.g., to 128×128). In these cases, models tend to oversmooth outputs or hallucinate details.

This degradation reflects physical reality: As resolution increases, unresolved subgrid physics (e.g., turbulence, stratification, and nonlinear instabilities) becomes dominant. The coarse input data no longer contain sufficient information to accurately infer high-resolution dynamics.

\subsection*{Future Work}

The previously discussed downscaling limitations of our models suggest two immediate directions for future research. One direction is to further improve the downscaling capability of both the PDE surrogate models and the static downscaling models applied to Copernicus sea velocity data. Another important avenue is to theoretically characterize the limits of these models, particularly as the governing physical behavior and equations change with increasing resolution. Incorporating uncertainty quantification through probabilistic neural operators and ensemble-based diffusion strategies could also help express confidence in high-resolution outputs, especially in underdetermined or data-sparse regimes.

\section{Visualizations}
\label{app:visualizations}
Here we provide some visualization for the downscaling PDE solver and the benchmarking of DFNO variants on Copernicus data.
\subsection{Downscaling and predicting PDE solutions}
\begin{figure}[h]
    \centering
    \includegraphics[width=0.9\linewidth]{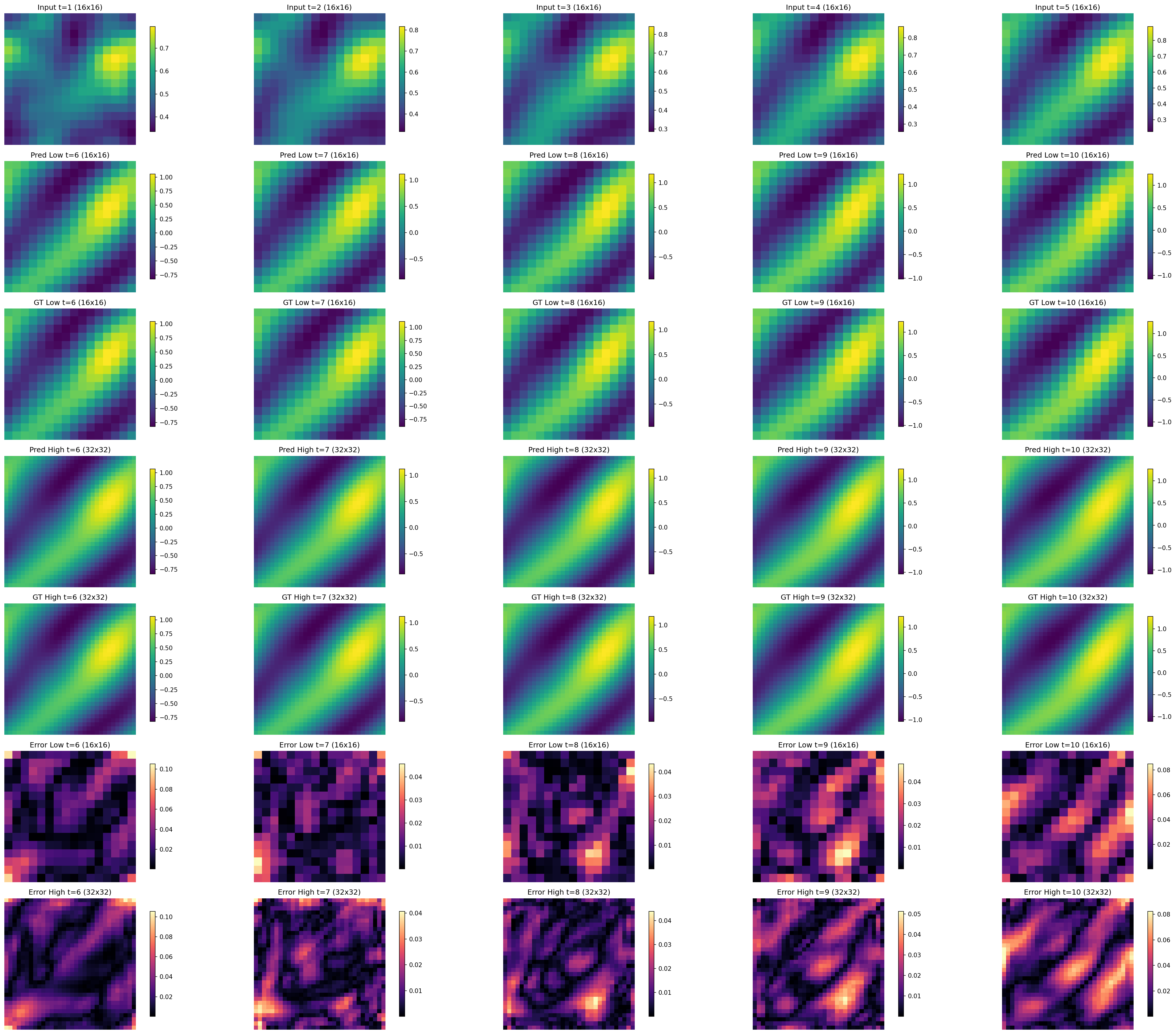}
    \caption{(Temp\_DFNO) 5 steps low resolution inputs, and predictions of the model on both $16\times16$ and $32\times32$ resolutions, as well as the residuals in the last 2 rows.}
    \label{fig:enter-label}
\end{figure}

\begin{figure}[h]
    \centering
    \includegraphics[width=\linewidth]{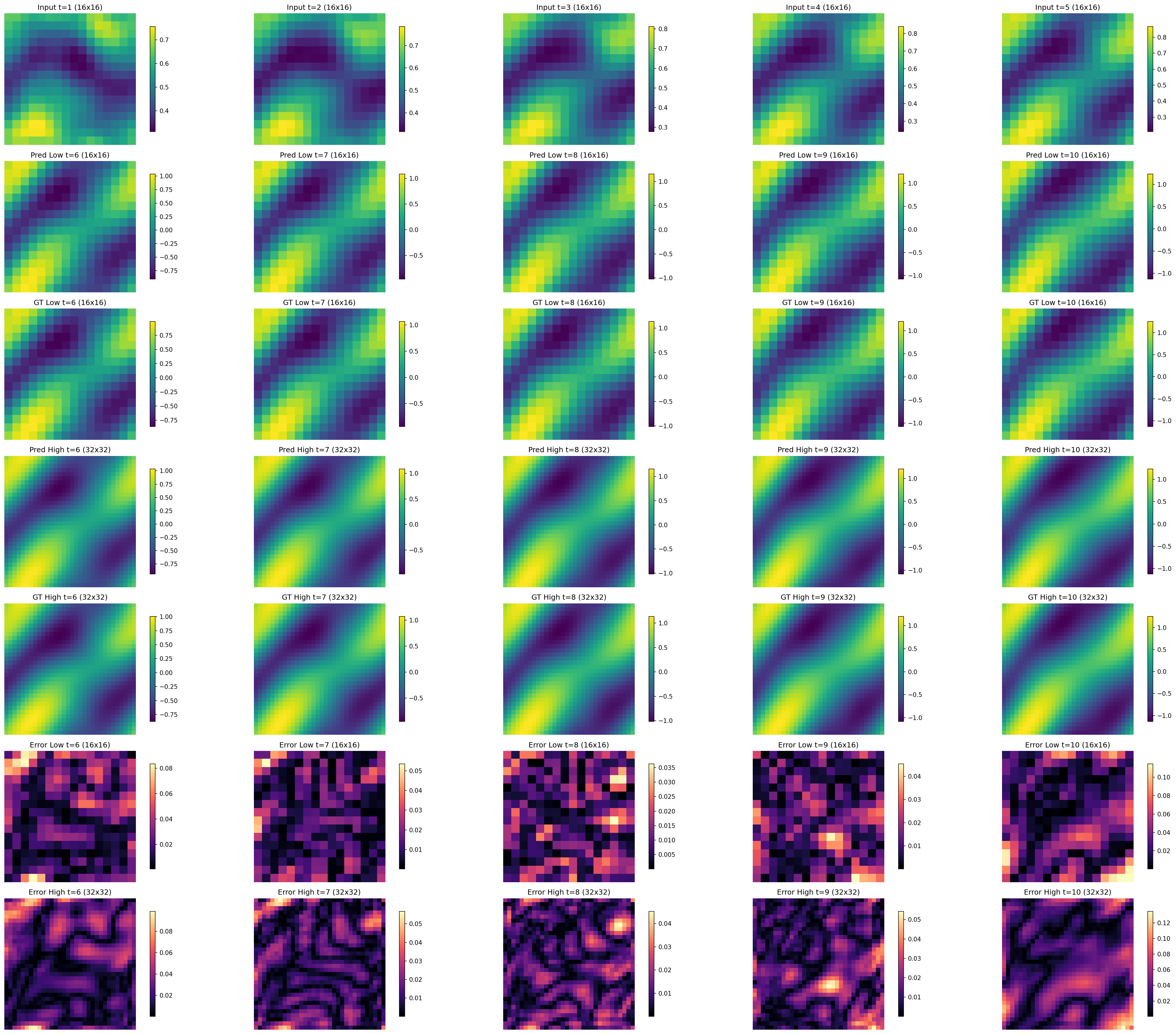}
    \caption{(Temp\_specDFNO) 5 steps low resolution inputs, and predictions of the model on both $16\times16$ and $32\times32$ resolutions, as well as the residuals in the last 2 rows. }
    \label{fig:figPDE}
\end{figure}
\clearpage
\subsection{Copernicus ocean current  Downscaling}
\begin{figure}[h]
    \centering
    \includegraphics[width=0.8\linewidth]{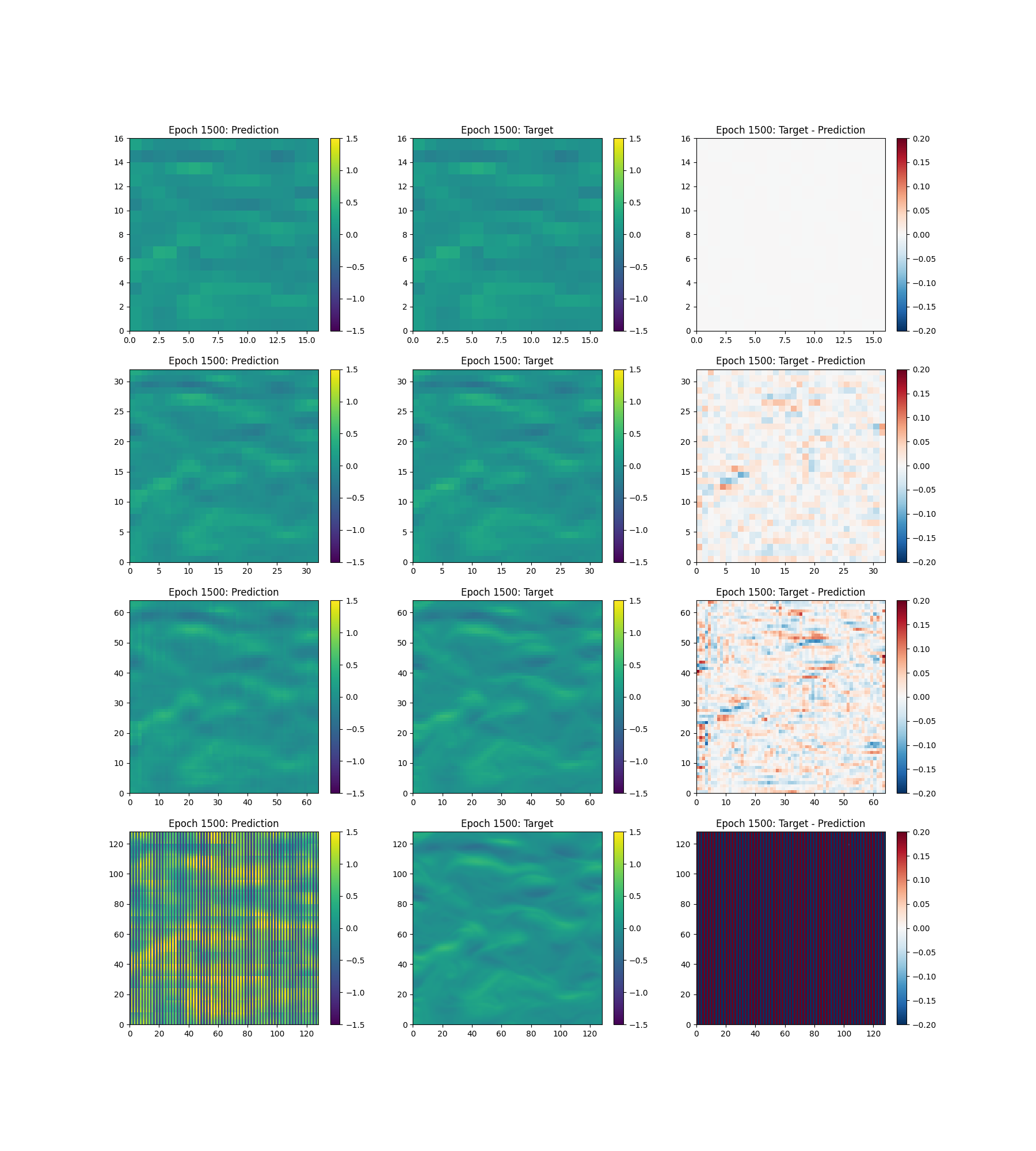}
    \caption{Ground truth vs. predictions of SpecDFNO. Rows correspond to different output resolutions: $16\times16$, $32\times32$, $64\times64$, and $128\times128$. The first column shows the model predictions, the second shows the ground truth, and the third displays the difference between them.}
    
    \label{fig:enter-label}
\end{figure}

\begin{figure}[h]
    \centering
    \includegraphics[width=\linewidth]{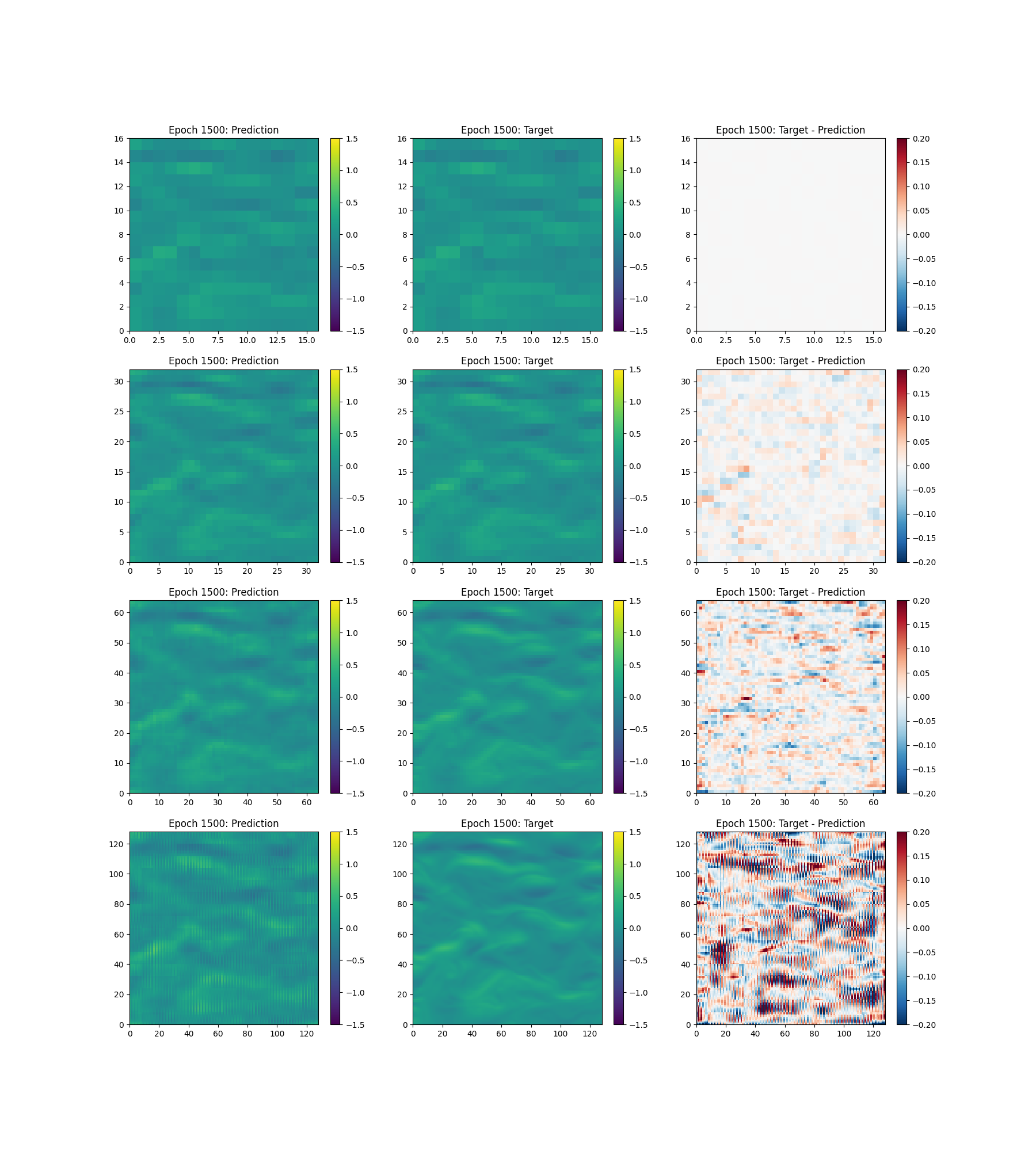}
    \caption{Ground truth vs. predictions of SpecDFNODiff. Rows correspond to different output resolutions: $16\times16$, $32\times32$, $64\times64$, and $128\times128$. The first column shows the model predictions, the second shows the ground truth, and the third displays the difference between them.}
    
    \label{fig:enter-label}
\end{figure}

\end{document}